\documentclass[conference]{IEEEtran}
\IEEEoverridecommandlockouts
\usepackage{cite}
\usepackage{amsmath,amssymb,amsfonts}
\usepackage{algorithm2e}
\usepackage{graphicx}
\usepackage{textcomp}
\usepackage{algpseudocode}
\usepackage{booktabs}
\usepackage{xcolor}
\usepackage{pifont}
\usepackage{placeins}
\def\BibTeX{{\rm B\kern-.05em{\sc i\kern-.025em b}\kern-.08em
    T\kern-.1667em\lower.7ex\hbox{E}\kern-.125emX}}
\begin{document}

\title{Investigating Imbalances Between SAR and Optical Utilization for Multi-Modal Urban Mapping\\
\thanks{This work was supported by the Swedish National Space Agency under Grant dnr 155/15; Digital Futures under the grant for the EO-AI4GlobalChange Project; the ESA-China Dragon 5 Program under the EO-AI4Urban Project; and the EU Horizon 2020 project HARMONIA under agreement No. 101003517.}
}

\author{\IEEEauthorblockN{Sebastian Hafner}
\IEEEauthorblockA{\textit{KTH Royal Institute of Technology}\\
Stockholm, Sweden \\
shafner@kth.se}
\and
\IEEEauthorblockN{Yifang Ban}
\IEEEauthorblockA{\textit{KTH Royal Institute of Technology}\\
Stockholm, Sweden \\
yifang@kth.se}
\and
\IEEEauthorblockN{Andrea Nascetti}
\IEEEauthorblockA{\textit{KTH Royal Institute of Technology}\\
Stockholm, Sweden \\
nascetti@kth.se}
}

\maketitle

\begin{abstract}
Accurate urban maps provide essential information to support sustainable urban development. Recent urban mapping methods use multi-modal deep neural networks to fuse Synthetic Aperture Radar (SAR) and optical data. However, multi-modal networks may rely on just one modality due to the greedy nature of learning. In turn, the imbalanced utilization of modalities can negatively affect the generalization ability of a network. In this paper, we investigate the utilization of SAR and optical data for urban mapping. To that end, a dual-branch network architecture using intermediate fusion modules to share information between the uni-modal branches is utilized. A cut-off mechanism in the fusion modules enables the stopping of information flow between the branches, which is used to estimate the network's dependence on SAR and optical data. While our experiments on the SEN12 Global Urban Mapping dataset show that good performance can be achieved with conventional SAR-optical data fusion (F1 score = 0.682 $\pm$ 0.014), we also observed a clear under-utilization of optical data. Therefore, future work is required to investigate whether a more balanced utilization of SAR and optical data can lead to performance improvements.
\end{abstract}


\section{Introduction}

The extent of urban areas is an important indicator of urbanization. The development of accurate and robust urban mapping methods is, therefore, essential to support sustainable urban development. Urban mapping methods are typically based on remotely sensed data due to their capability to consistently cover large geographical areas. In particular, satellite data from Synthetic Aperture Radar (SAR) sensors have been proven to be an invaluable data source for urban mapping (e.g., \cite{ban2015spaceborne}). Build-up areas in SAR imagery are characterized by high backscattering due to the double bounce effect of buildings. Apart from SAR data, the success of computer vision techniques such as convolutional neural networks has led to the development of promising mapping methods using optical satellite imagery (e.g., \cite{qiu2020framework, corbane2021convolutional}).

Recently, several studies investigated multi-modal learning with SAR-optical data fusion for urban mapping \cite{hafner2021exploring, hafner2022unsupervised, schmitt2020weakly} and urban change detection \cite{hafner2021sentinel, ebel2021fusing, hafner2022multi}. These works performed multi-modal learning by combining Sentinel-1 SAR and Sentinel-2 MultiSpectral Instrument (MSI) data with either input-level or decision-level fusion using dual branch architectures. The aforementioned works, therefore, rely on multi-modal networks to effectively use the additional information in the form of other modalities to improve model performance upon uni-modal networks. However, some studies reported unsatisfactory performance of SAR-optical data fusion \cite{schmitt2020weakly, ebel2021fusing}. At the root of the problem could be the under-utilization of either modality. In fact, Wu \textit{et al.} \cite{wu2022characterizing} recently demonstrated that multi-modal learning processes for many domains are predominately relying on the modality that is the faster to learn from. They refer to this as the greedy nature of multi-modal learning, which can hamper the generalization ability of models \cite{wu2022characterizing}. For urban mapping with SAR-optical data fusion, this may be a crucial issue, since the dependence on a single modality has been demonstrated to be insufficient for urban mapping at a global scale \cite{hafner2022unsupervised}.

In this paper, we aim to uncover imbalances between SAR and optical modality utilization for urban mapping. Specifically, we apply recently published concepts to characterize the greedy nature of multi-modal learning (i.e.,  \cite{wu2022characterizing}) to an urban mapping dataset featuring Sentinel-1 SAR and Sentinel-2 data. Our investigation of a multi-modal network's dependence on SAR and optical data may contribute to the development of models with better generalization ability in the future.

\section{Dataset}

\begin{figure}[h]
    \centering
    \includegraphics[width=.48\textwidth]{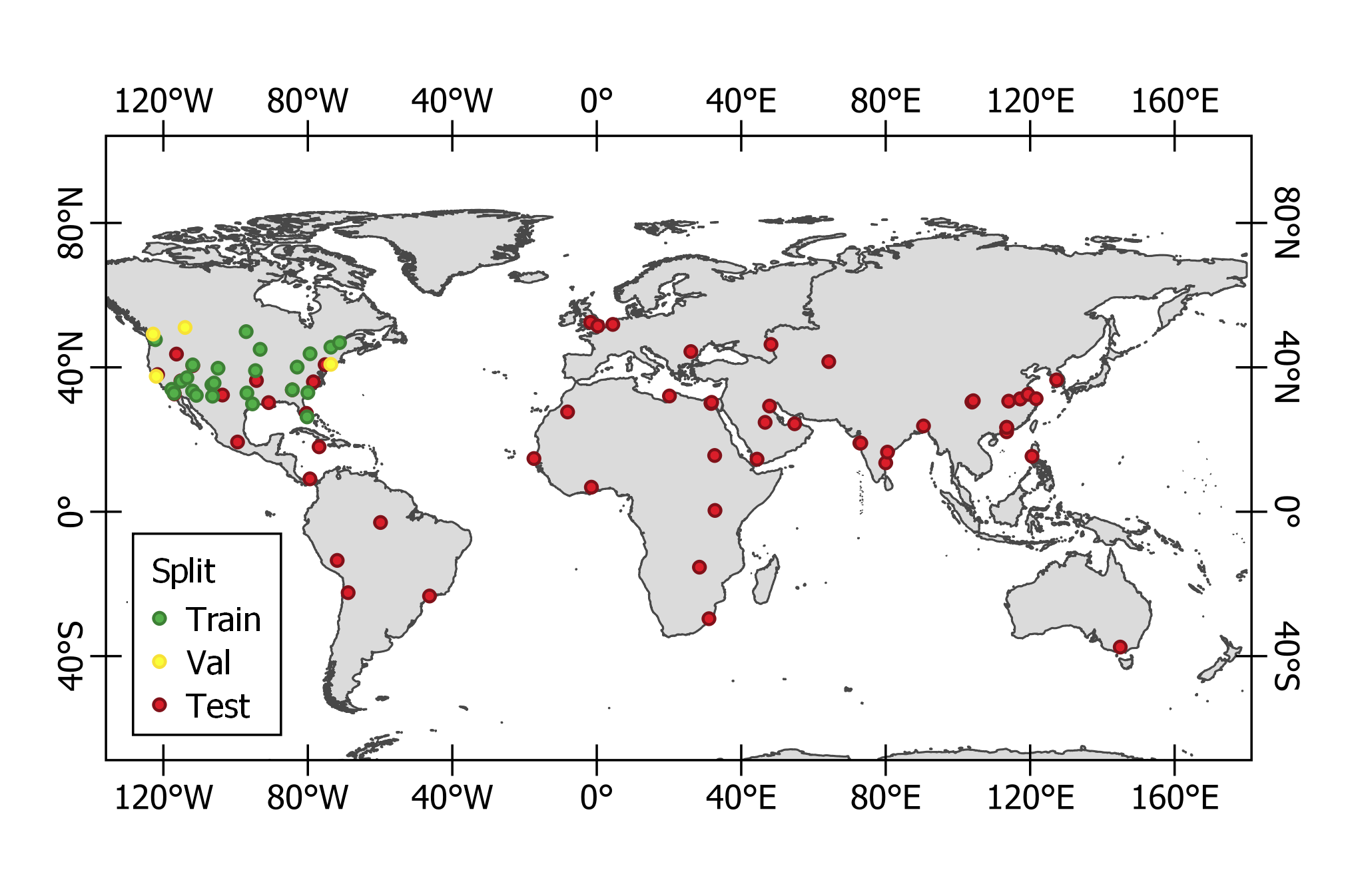}
    \caption{Locations of the training, validation, and test sites of the SEN12 Global Urban Mapping dataset \cite{hafner2022unsupervised}}.
    \label{fig:study_area}
\end{figure}

We consider the urban mapping problem with multi-modal satellite data posed by the SEN12 Global Urban Mapping (SEN12\_GUM) dataset\footnote{https://doi.org/10.5281/zenodo.6914898} \cite{hafner2022unsupervised}. The SEN12\_GUM dataset, denoted by $\mathcal{D}$, consists of multiple instances of mean Sentinel-1 SAR images, $x_{\rm sar}$, and median Sentinel-2 MSI images, $x_{\rm opt}$, acquired over the same area. In addition, corresponding urban label images, $y \in \{0, 1\}$, are provided. The dataset is partitioned into a training, validation and test set, denoted by $\mathcal{D}^{\rm train}$, $\mathcal{D}^{\rm val}$ and $\mathcal{D}^{\rm test}$, respectively. The 26 training and 4 validation sites are located in the United States, Canada and Australia. On the other hand, the 60 test sites are globally distributed and cover unique geographies to test the generalization ability of networks. An overview of the locations of the sites is given in Figure \ref{fig:study_area}.

\section{Methodology}

\subsection{Multi-Modal Deep Learning Model}

\begin{figure}[h]
    \centering
    \includegraphics[width=.48\textwidth]{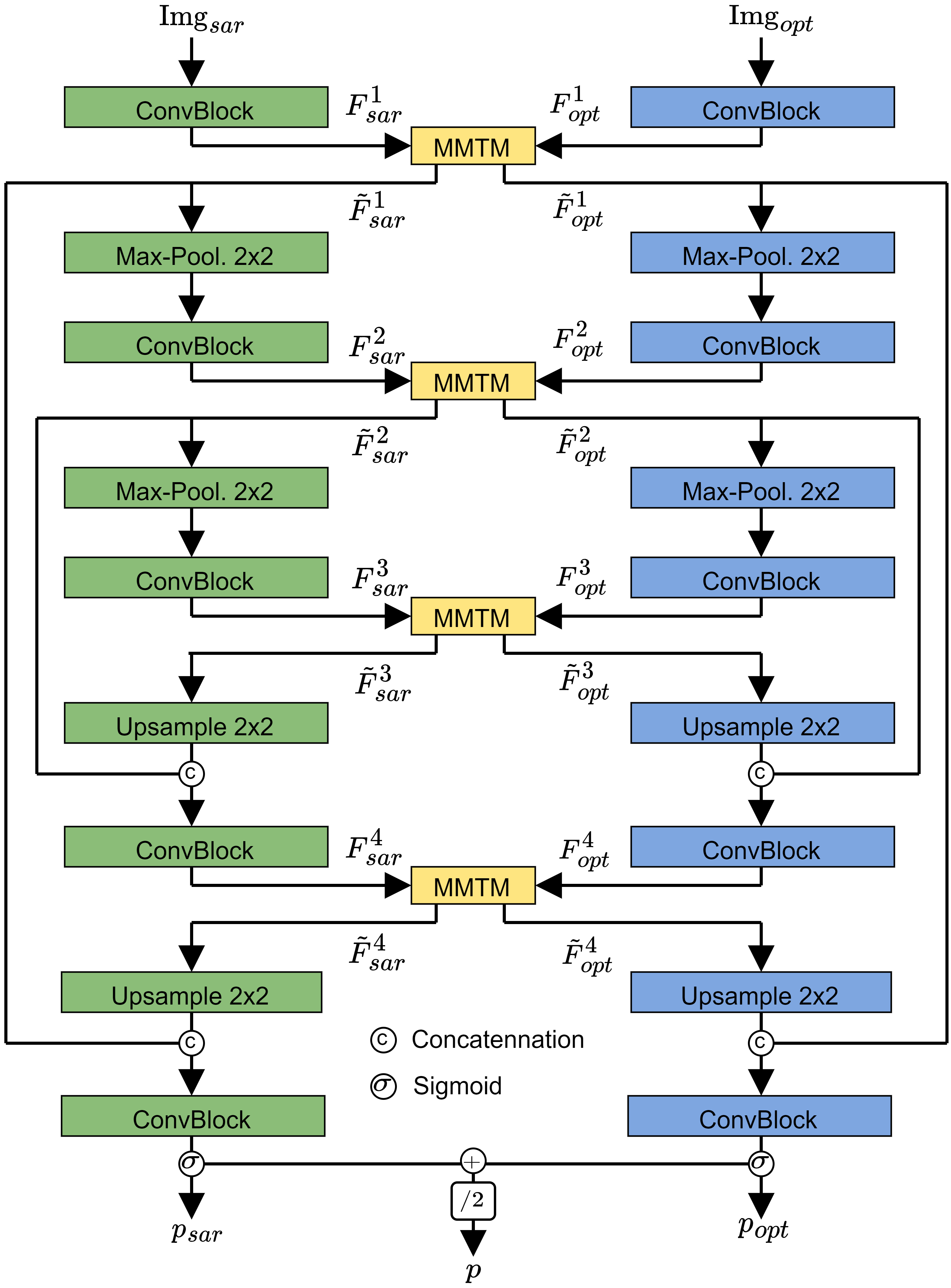}
    \caption{Overview of the architecture consisting of two uni-modal U-Nets \cite{ronneberger2015u} that are connected through four Multi-Modal Transfer Modules (MMTMs) \cite{joze2020mmtm} for feature modality fusion in convolution layers.}
    \label{fig:mmtm_dsunet}
\end{figure}

We design a multi-modal network, $f(.)$, consisting of two identical uni-modal branches, $f_{\rm sar}(.)$ and $f_{\rm opt}(.)$, each following the architecture of U-Net \cite{ronneberger2015u} (Figure \ref{fig:mmtm_dsunet}). To enable the flow of modality-specific information between convolution layers of the uni-modal U-Net branches, four feature fusion modules, specifically Multi-Modal Transfer Modules (MMTMs) \cite{joze2020mmtm}, are incorporated into the network. MMTMs first squeeze spatial information from the uni-modal feature tensors, $F_{\rm sar}$ and $F_{\rm opt}$, into respective vectors, $h_{\rm sar}$ and $h_{\rm opt}$, using global average pooling operations. A joint representation, $Z$, is then predicted from the concatenated vectors using a fully-connected layer. The joint representation is converted to two excitation signals, $E_{\rm sar}$ and $E_{\rm opt}$, using two additional fully-connected layers. Finally, the resulting excitation signals are used to recalibrate the uni-modal feature tensors using a simple gating mechanism. Following that, the MMTM cross-modal recalibration generates two outputs, $\tilde{F}_{sar}$ and $\tilde{F}_{opt}$, that are passed back to the uni-modal branches. In the end, each uni-modal branch produces a separate prediction for an input $(x_{\rm sar}, x_{\rm opt})$: 

\begin{equation}
    p_{\rm sar} = f_{\rm sar}(x_{\rm sar}, x_{\rm opt}), \ p_{\rm opt} = f_{\rm opt}(x_{\rm sar}, x_{\rm opt}),
\end{equation}

Finally, the predictions of the branches are combined to form the fusion output of the multi-modal network:

\begin{equation}
    p = \frac{1}{2}(p_{\rm sar} + p_{\rm opt}).
\end{equation}

The network is trained with two modality-specific losses that are directly applied to the predictions of the uni-modal branches, i.e., the loss is $\mathcal{L}(y, p_{\rm sar}) + \mathcal{L}(y, p_{\rm opt})$, where $y$ denotes the label.

An important characteristic of the multi-modal network is that the connection to share information between the uni-modal branches can be cut off to produce predictions that rely on only one of the two modalities:

\begin{equation}
    p'_{\rm sar} = f'_{\rm sar}(x_{\rm sar}), \: p'_{\rm opt} = f'_{\rm opt}(x_{\rm opt}).
\end{equation}

The cut-off mechanism is implemented in the MMTM by approximating the vectors $h_{\rm sar}$ and $h_{\rm opt}$ based on the training set. Specifically, $p'_{\rm sar}$ is obtained by replacing the information coming from the optical branch, i.e., $h_{\rm opt}$, with the average of $h_{\rm opt}$ over $\mathcal{D}^{\rm train}$, and vice versa $h_{\rm sar}$ is replaced with the average of $h_{\rm sar}$ over $\mathcal{D}^{\rm train}$ for $p'_{\rm opt}$. The average values of $h_{\rm sar}$ and $h_{\rm opt}$ are defined as follows:

\begin{equation}
\overline{h}_{\rm sar} = \frac{1}{n}\sum_{i=1}^n h_{\rm sar}(x^i), \ \overline{h}_{\rm opt} = \frac{1}{n}\sum_{i=1}^n h_{\rm opt}(x^i),
\end{equation}

where $n$ is the number of samples in $\mathcal{D}_{\rm train}$ and $h_{\rm sar}(x^i)$ and $h_{\rm sar}(x^i)$ represent the value of $h_{\rm sar}$ and $h_{\rm opt}$ with the $i^{\rm th}$ sample, $x^i$ as input.

\subsection{Measuring Imbalance between Modality Utilization}

The concept of Conditional Utilization Rates (CURs) was recently introduced by Wu \textit{et al.} \cite{wu2022characterizing} with the goal to analyse a multi-modal network's ability to utilize both modalities. In more detail, a multi-modal network’s CUR for a given modality is the relative difference in accuracy between two sub-networks derived from the network, one using both modalities and the other using only one modality. The CURs for SAR and optical data are defined as follows:

\begin{equation}
    u(sar|opt) = \frac{A(y, p_{\rm opt})-A(y, p'_{\rm opt})}{A(y, p_{\rm opt})},
\end{equation}

and

\begin{equation}
    u(opt|sar) = \frac{A(y, p_{\rm sar})-A(y, p'_{\rm sar})}{A(y, p_{\rm sar})},
\end{equation}

where $A(y, p)$ denotes the accuracy value obtained from prediction $p$ with label $y$. Consequently, $A(y, p)$ and $A(y, p')$ correspond to the accuracy values obtained from networks using both modalities and only one modality, respectively.

The CUR of SAR given optical, denoted by $u(sar|opt)$, measures, for example, how important it is for the model to use SAR data in order to reach accurate predictions, given the presence of optical data. In other words, $u(sar|opt)$ is the marginal contribution of $x_{\rm sar}$ to increasing the accuracy of $p_{opt}$. For this case, it is generally assumed that $A(y, p_{\rm opt}) > A(y, p'_{\rm opt})$, since $p_{\rm opt}$ was obtained from the combination of $x_{\rm sar}$ and $x_{\rm opt}$, while $p'_{\rm opt}$ was obtained from $x_{\rm opt}$ alone.

The difference between CURs is used to measure the imbalance between the utilization of SAR and optical data:

\begin{equation}
    d_{\rm util} = u(sar|opt) - u(opt|sar).
\end{equation}

Since CURs are assumed to be positive, $d_{\rm util}$ is bound by the range $[-1, 1]$. When $d_{\rm util}$ is close to either bound, the network's ability to accurately predict $p_{\rm sar}$ and $p_{\rm opt}$ comes only from one of the modalities. Thus, high $|d_{\rm util}|$ values imply an imbalance between the utilization of modalities.

\subsection{Experimental Setup}

We train 5 networks with different seeds for weight initialization, dataset shuffling, and data augmentation. Networks are trained for 15 epochs with a batch size of 8, and AdamW \cite{loshchilov2017decoupled} is used as optimizer with an initial learning rate of $10^{-4}$. Flips (horizontal and vertical) and rotations ($k * 90^{\circ}$, where $k \in \{0, 1, 2, 3\}$) are used as data augmentations. As loss function, a Jaccard-like loss, specifically the Power Jaccard loss \cite{duque2021power}, is used. As mentioned earlier, the loss is separately applied to the predictions of the uni-modal branches of the multi-modal network. 

To measure the accuracy of predictions, the commonly used accuracy metric F1 score, Precision (P), and Recall (R) are employed. They are defined as follows:

\begin{equation}
\label{eq:f1_score}
\begin{aligned}
    P = \frac{TP}{TP + FP} \:
    R = \frac{TP}{TP + FN} \:
    F1 = 2 * \frac{P * R}{P + R}
\end{aligned}
\end{equation}

where TP, FP, and FN represent the number of true positive, false positive, and false negative pixels, respectively.

\section{Results}

Table \ref{tab:quantitative_results} lists network performance on the test set for the three network outputs, namely SAR ($p_{\rm sar}$), optical ($p_{\rm opt}$) and fusion ($p$). For all three accuracy metrics, the fusion output achieved the highest mean value. With a mean F1 score of 0.682, the fusion prediction of the network is, however, only slightly more accurate than the optical prediction (0.673). Among the uni-modal branches, the optical branch $f_{\rm opt}(.)$ performed notably better than the SAR branch $f_{\rm sar}(.)$. In particular, the detection rate of urban areas indicated by mean recall, is considerably higher for optical (0.621) in comparison to SAR (0.557).

\begin{table}[h]
\small
  \caption{Urban mapping performances on the test set. Values correspond to the mean $\pm$ 1 standard deviation of 5 runs.}
  \label{tab:quantitative_results}
  \centering
  \begin{tabular}{lccc}
    \toprule
    Prediction & F1 score & Precision & Recall \\
    \midrule
    SAR & 0.613 $\pm$ 0.006 & 0.683 $\pm$ 0.032 & 0.557 $\pm$ 0.014 \\
    Optical & 0.673 $\pm$ 0.013 & 0.734 $\pm$ 0.031 & 0.621 $\pm$ 0.009 \\
    Fusion & \textbf{0.682} $\pm$ \textbf{0.014} & \textbf{0.746} $\pm$ \textbf{0.031} & \textbf{0.630} $\pm$ \textbf{0.013} \\
    \bottomrule
  \end{tabular}
\end{table}

In terms of modality utilization, CURs of 0.37 ($\pm$ 0.12) and 0.17 ($\pm$ 0.04) were recorded for $u(sar|opt)$ and $u(opt|sar)$, respectively. Consequently, $|d_{util}|$, the absolute difference between CURs, is 0.20 ($\pm$ 0.14). These results indicate that the addition of SAR data given optical data considerably increased the accuracy of the prediction of the optical branch $f_{\rm opt}$; in contrast, the increase in accuracy from the addition of optical data given SAR data is smaller. Thus, an imbalance between the utilization of SAR and optical data exists, as also indicated by $|d_{\rm util}|$.

\begin{figure*}[h]
    \centering
    \includegraphics[width=\textwidth]{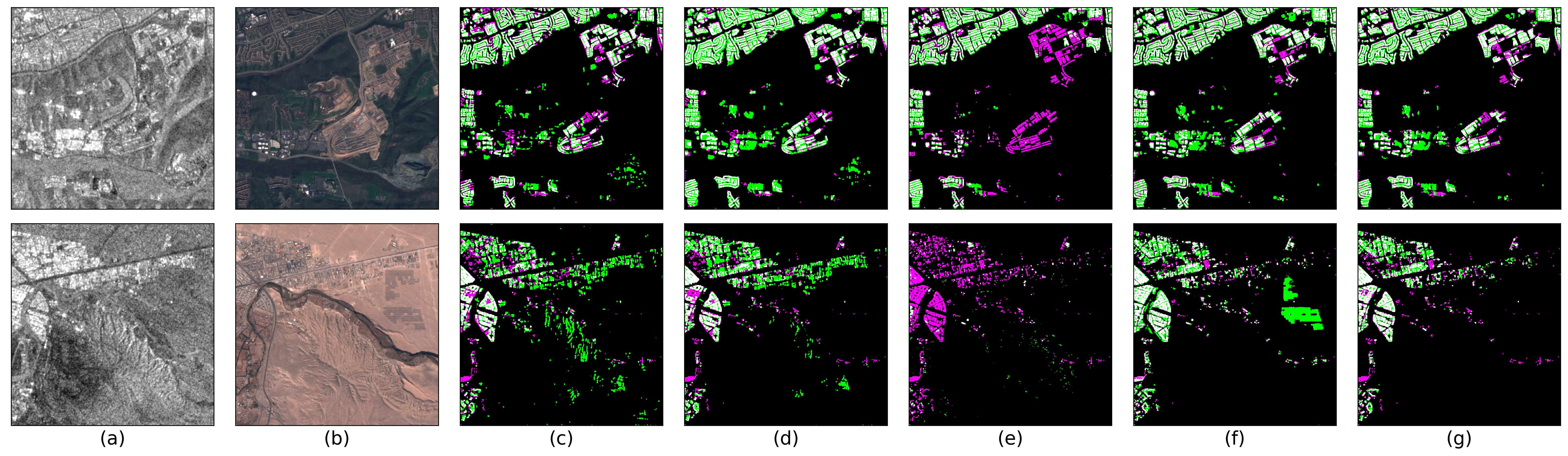}
    \caption{Qualitative results for two test sites (rows). Column (a) shows the Sentinel-1 SAR VV band and column (b) the true color (red: B4, green: B3, blue: B2) Sentinel-2 MSI image. Columns (c-g) show network predictions, where true positives, true negatives, false positives, and false negatives are visualized in white, black, green, and magenta, respectively. The SAR branch predictions with cross-modal flow turned off and on are shown in columns (c) and (d), respectively. Likewise, the optical branch predictions are shown in columns (e) and (f). Finally, column (d) shows the fused prediction of (d) and (f).}
    \label{fig:qualitative_results}
\end{figure*}

Figure \ref{fig:qualitative_results} shows the qualitative test results for two sites (rows). The results were produced with the network yielding the median accuracy in terms of F1 score. Generally, it is apparent that the SAR predictions (c \& d) are similar, independent of whether the flow of information between the uni-modal branches is on (c) or off (d). This indicates that enabling the flow of optical data to the SAR branch has little effect on the results, which is supported by the network's low CUR of optical data given SAR data (0.19). In contrast, stark differences are apparent between the predictions of the optical branch (e \& f). In particular, the recalibration of the optical features with SAR information helped to reduce false negatives (magenta) for all sites. These observations are in line with the CUR of SAR given optical (0.41)  However, it can also have negative effects, as exemplified by the green areas indicating false positives for the site on the second row. The fusion output (g), obtained from the outputs of the uni-modal branches (d \& f), generally delineates urban areas the most accurately among the other network outputs.

\section{Discussion}

We find MMTMs to be an effective tool for intermediate data fusion as part of the multi-modal network architecture. In fact, with a mean F1 score of 0.682, the network outperforms not only the baseline uni-modal SAR and optical approach on the SEN12\_GUM dataset (0.574 and 0.580, respectively), but also the input-level fusion one (0.651) \cite{hafner2022unsupervised}. The network was only outperformed by the unsupervised domain adaptation approach in \cite{hafner2022unsupervised} (F1 score = 0.697); however, the domain adaptation approach exploited additional unlabeled training data to adapt the model to the test set, while the network in this paper is solely relying on labeled data.

In terms of modality utilization, we observed an imbalance between the utilization of SAR and optical data for urban mapping. Specifically, while SAR data contributed to increasing the accuracy of the predictions of the optical branch (mean $u(sar|opt)$ = 0.37), optical data was strongly under-utilized by the SAR branch (mean $u(opt|sar)$ = 0.17). Thus, there is an imbalance in modality utilization. Wu \textit{et al.} \cite{wu2022characterizing} hypothesized that multi-modal networks primarily rely on the modality that is the fastest to learn from. Since our networks showed a strong dependence on SAR data, we assume that the simple representation of built-up areas in SAR imagery (i.e., high backscattering) facilitates learning from SAR data compared to optical data where a complex spectral heterogeneity exists for urban landscapes. Our findings, therefore, indicate that urban mapping from SAR and optical data requires an incentive to utilize optical data in order to balance the multi-modal learning process.

\section{Conclusion}

In this paper, we employed a dual-branch architecture with MMTM modules to investigate imbalances between SAR and optical utilization for multi-modal urban mapping. Our experiments on the SEN12\_GUM dataset demonstrated that the architecture is effective for urban mapping. However, this work also demonstrated that the multi-modal networks trained on SAR and optical data rely more on the former than on the latter modality. Consequently, optical data is under-utilized.

Future work will investigate methods to balance the learning from SAR and optical data. In particular, the balanced multi-modal learning algorithm proposed in \cite{wu2022characterizing} will be investigated for urban mapping from SAR and optical data.

\bibliography{ref.bib}

\begin{thebibliography}{10}

\bibitem{ban2015spaceborne}
Ban, Y. et~al,
\newblock ``Spaceborne sar data for global urban mapping at 30 m resolution
  using a robust urban extractor,''
\newblock {\em ISPRS Journal of Photogrammetry and Remote Sensing}, vol. 103,
  pp. 28--37, 2015.

\bibitem{qiu2020framework}
Qiu, C. et~al,
\newblock ``A framework for large-scale mapping of human settlement extent from
  sentinel-2 images via fully convolutional neural networks,''
\newblock {\em ISPRS Journal of Photogrammetry and Remote Sensing}, vol. 163,
  pp. 152--170, 2020.

\bibitem{corbane2021convolutional}
Corbane, C. et~al,
\newblock ``Convolutional neural networks for global human settlements mapping
  from sentinel-2 satellite imagery,''
\newblock {\em Neural Computing and Applications}, vol. 33, no. 12, pp.
  6697--6720, 2021.

\bibitem{hafner2021exploring}
Hafner, S. et~al,
\newblock ``Exploring the fusion of sentinel-1 sar and sentinel-2 msi data for
  built-up area mapping using deep learning,''
\newblock in {\em 2021 IEEE International Geoscience and Remote Sensing
  Symposium IGARSS}. IEEE, 2021, pp. 4720--4723.

\bibitem{hafner2022unsupervised}
Hafner, S. et~al,
\newblock ``Unsupervised domain adaptation for global urban extraction using
  sentinel-1 sar and sentinel-2 msi data,''
\newblock {\em Remote Sensing of Environment}, vol. 280, pp. 113192, 2022.

\bibitem{schmitt2020weakly}
Schmitt, M. et~al,
\newblock ``Weakly supervised semantic segmentation of satellite images for
  land cover mapping--challenges and opportunities,''
\newblock {\em arXiv preprint arXiv:2002.08254}, 2020.

\bibitem{hafner2021sentinel}
Hafner, S. et~al,
\newblock ``Sentinel-1 and sentinel-2 data fusion for urban change detection
  using a dual stream u-net,''
\newblock {\em IEEE Geoscience and Remote Sensing Letters}, vol. 19, pp. 1--5,
  2021.

\bibitem{ebel2021fusing}
Ebel, P. et~al,
\newblock ``Fusing multi-modal data for supervised change detection,''
\newblock {\em The International Archives of the Photogrammetry, Remote Sensing
  and Spatial Information Sciences}, vol. 43, pp. 243--249, 2021.

\bibitem{hafner2022multi}
Hafner, S.,
\newblock {\em Multi-Modal Deep Learning with Sentinel-1 and Sentinel-2 Data
  for Urban Mapping and Change Detection},
\newblock Ph.D. thesis, KTH Royal Institute of Technology, 2022.

\bibitem{wu2022characterizing}
Wu, N. et~al,
\newblock ``Characterizing and overcoming the greedy nature of learning in
  multi-modal deep neural networks,''
\newblock in {\em International Conference on Machine Learning}. PMLR, 2022,
  pp. 24043--24055.

\bibitem{ronneberger2015u}
Ronneberger, O. et~al,
\newblock ``U-net: Convolutional networks for biomedical image segmentation,''
\newblock in {\em International Conference on Medical image computing and
  computer-assisted intervention}. Springer, 2015, pp. 234--241.

\bibitem{joze2020mmtm}
Joze, H.R.V. et~al,
\newblock ``Mmtm: Multimodal transfer module for cnn fusion,''
\newblock in {\em Proceedings of the IEEE/CVF Conference on Computer Vision and
  Pattern Recognition}, 2020, pp. 13289--13299.

\bibitem{loshchilov2017decoupled}
Loshchilov, I. and Hutter, F.,
\newblock ``Decoupled weight decay regularization,''
\newblock {\em arXiv preprint arXiv:1711.05101}, 2017.

\bibitem{duque2021power}
Duque-Arias, D. et~al,
\newblock ``On power jaccard losses for semantic segmentation,''
\newblock in {\em VISAPP 2021: 16th International Conference on Computer Vision
  Theory and Applications}, 2021.

\end{thebibliography}
\bibliographystyle{IEEEbib}

\end{document}